%% file: main.tex
\documentclass[lettersize,journal]{IEEEtran}
\pdfoutput=1

\usepackage{amsmath,amsfonts,amssymb}
\usepackage{array}
\usepackage{adjustbox}
\usepackage{chapterbib}
\usepackage{balance}
\usepackage{cite}
\usepackage{enumerate}
\usepackage{graphicx}
\usepackage{gensymb}
\usepackage{multirow}
\usepackage{stfloats}
\usepackage{hyperref}
\usepackage{soul}
\usepackage{tabularx}
\usepackage{textcomp}
\usepackage[caption=false,font=normalsize,labelfont=sf,textfont=sf]{subfig}
\usepackage{url}
\usepackage{verbatim}
\usepackage{xcolor}
\definecolor{ninetyfive}{rgb}{0.89, 0.1, 0.1}
\definecolor{ninetynine}{rgb}{0.89, 0.1, 0.1}
\definecolor{noeffect}{rgb}{0.0, 0.0, 0.0}

\newcommand{\ninety}[0]{Perpendicular}
\newcommand{\adjacent}[0]{Adjacent}
\newcommand{\opposite}[0]{Opposite}
\newcommand{\linear}[0]{Linear }

\definecolor{blizzardblue}{rgb}{0, 0.53, 0.91}
\definecolor{blizzardred}{rgb}{0.91, 0.53, 0}
\definecolor{junglegreen}{rgb}{0.16, 0.67, 0.52}

\newcommand{\newtext}[1]{\textcolor{black}{#1}}

\title{\newtext{A Study in Zucker: Insights on Interactions\\ Between Humans and Small Service Robots}}

\author{Alex Day and Ioannis Karamouzas%
\thanks{This work was supported by the NSF under Grant No. IIS-2047632.}
\thanks{Alex Day is with the School of Computing at Clemson University, SC, USA (email: \texttt{adday@clemson.edu}). Ioannis Karamouzas is with the Department of Computer Science and Engineering at University of California, Riverside, USA, and a visiting professor at CYENS - Centre of Excellence, Nicosia, Cyprus (email: \texttt{ioannis@cs.ucr.edu}).  This work was initiated when he was at Clemson University.}
}

\begin{document}

\include{text}
\include{appendix}

\end{document}

%% file: text.tex
{\maketitle}

\begin{abstract}
Despite recent advancements in human-robot interaction (HRI), there is still limited knowledge about how humans interact and behave in the presence of small service indoor robots and, subsequently, about the human-centered behavior of such robots. This also raises concerns about the applicability of current trajectory prediction methods to indoor HRI settings as well as the accuracy of existing crowd simulation models in shared environments. To address these issues, we introduce a new HRI dataset focusing on interactions between humans and small differential drive robots running different types of controllers. Our analysis shows that anticipatory and non-anticipatory robot controllers impose similar constraints to humans’ safety and efficiency. Additionally, we found that current state-of-the-art models for human trajectory prediction can adequately extend to indoor HRI settings. Finally, we show that humans respond differently to small differential drives than to other humans when collisions are imminent, since interacting with small robots can only cause a finite level of social discomfort as compared to human-human interactions. Our dataset and related code and models are available at: https://motion-lab.github.io/ZuckerDataset.
\end{abstract}


\section{Introduction}
Given the advancements in robotics and AI in recent years along with the significant decrease in hardware cost,  robots are increasingly becoming a part of our lives, from vacuuming our floors and delivering items in automated warehouses to transporting passengers autonomously in urban areas~\cite{eliot}. From robot vacuums to robo-taxis, studying how humans react to such robots and interact in a shared environment plays a crucial role in the further development of autonomous systems. In this paper, we focus on small service robots like Roombas that operate in indoor spaces and have to interact intelligently with humans to avoid collisions while completing their tasks.  In this field, there has been a lot of recent work on \emph{social robot navigation} aiming to steer robots in human-populated spaces~\cite{mavrogiannis2023core}. Despite such efforts, existing approaches are typically evaluated from a robot-centric perspective. While the robot's performance is important to assess the quality of employed navigation algorithms, the efficiency and safety of the humans are equally important in shared interaction settings. Unfortunately, in such settings, there is still limited knowledge about how humans interact and behave in the presence of robots and, subsequently, about the \emph{human-centered} behavior of robots.  This also raises concerns about the applicability of current trajectory prediction methods to indoor HRI settings as well as the accuracy of existing human simulation models.To address this gap in the literature we conducted a user study in the Zucker Family Graduate Education Center in Charleston, SC, USA. Following a brief overview of highly relevant work, we focus on the related results from our analysis. Our findings and the collected  HRI trajectories, which we share with the community, can enable more socially-aware robot control through providing a better understanding of the human-centered performance of different classes of robot controllers, facilitating more accurate human trajectory prediction in shared interaction settings, and informing the development of new pedestrian simulation methods.

\vspace{-3px}
\section{Related Work}
\subsection{HRI Datasets and Studies}
A lot of recent work has been focusing on capturing and analyzing HRI data to better understand how humans interact with robots. The related datasets can be classified based on if the interactions they contain are unstructured or structured. Unstructured interactions can closely approximate true-to-life behavior, but require laborious manual labeling due to the lack of control in the experimental conditions~\cite{zhimon2020jist,martin2021jrdb,karnan2022socially}. In contrast, structured interactions occur in a controlled environment, typically equipped with sophisticated motion capture solutions that facilitate high-fidelity data collection and analysis. The TH\"{O}R dataset \cite{thorDataset2018} is one such example capturing diverse indoor interactions between humans and between humans and a socially unaware small industrial robot following a predetermined path. In recent years, a number of HRI studies were conducted in structured settings to validate new methods for social navigation~\cite{kim2016socially,truong2017toward,kretzschmar2016socially}. More recently,  Lo et al.~\cite{lo2019perception} evaluate the effect that different navigation strategies of a self-balancing robot have on humans in a perpendicular crossing scenario.  Similarly, the HRI datasets in~\cite{zhang2022hri} and~\cite{chen2018pedestrian} study the impact that a robot has on humans during a crossing-gate and a corridor-exit experiment, respectively,  with~\cite{zhang2022hri} showing that participants behave more conservatively in the presence of a robotic wheelchair than a humanoid robot. More closely related to our work, the large-scale study of Mavrogiannis et al.~\cite{mavrogiannis2019effects, mavrogiannis2022social} evaluates different navigation algorithms through human-centered metrics including the energy expended by the humans and the smoothness of their paths.  Our work is complimentary to such aforementioned studies as we release a new structured HRI dataset containing a range of scenarios and differential drive controllers. Similar to~\cite{lo2019perception,mavrogiannis2022social} we seek to evaluate the behavior of different robot navigation methods through human-centered metrics. However, we propose an intuitive way to do so without  explicitly comparing the methods but by assessing the performance of the humans with and without the robot. Unique to prior work, we provide novel results  about the performance of human trajectory prediction methods in indoor HRI settings and quantify how humans resolve collisions with small service robots. 

\vspace{-10px}
\subsection{Social Robot Navigation Methods}
There has been a lot of highly relevant work for autonomous and collision-free navigation of a robot amidst a crowd of humans~\cite{mavrogiannis2023core}. Existing techniques for human-aware robot navigation can be broadly classified into geometric planning approaches and learning-based methods. State-of-the-art geometric planners rely on the concepts of velocity obstacles (VO) and time to collision~\cite{van2011reciprocal,karamouzas2014universal}, and can provide formal guarantees of the collision-free behavior of robots while supporting a wide range of robot models and behaviors~\cite{luo2018porca,davis2018nh}. However, they often require careful parameter tuning to achieve desired navigation results, with VO-based controllers also exhibiting  overly conservative behavior in an attempt to guarantee safe navigation~\cite{mavrogiannis2023core}. Learning-based methods can provide more flexibility,  with recent contributions taking advantage of deep reinforcement learning to enable end-to-end steering and train crowd-aware navigation policies~\cite{fan2020distributed,sathyamoorthy2020densecavoid,everett2021collision}. Despite the success stories of geometric and learning-based approaches the robots do not always exhibit the level of sophistication that humans do in similar interaction scenarios. As such, researchers have been focusing on improving the control policy of a robot by accounting for interactions between all agents in the scene, expert demonstrations, social norms, human collaboration, and group-aware planning among others~\cite{kim2016socially,kretzschmar2016socially,how,chen2019crowd,mavrogiannis2022winding,katyal2022learning,gonon2023inverse,bevilacqua2018reactive,liu2022intention}. In synergy with local navigation, research on HRI has also been focused on global planners to drive a mobile robot in a socially acceptable manner and facilitate human-like movements~\cite{kirby2010social,kruse2012legible,truong2017toward}.  In this paper, we are interested in assessing the performance of three representative robot navigation algorithms in indoor physical settings. While a lot of existing works in benchmarking social robot navigation methods typically employ a robot-centric evaluation paradigm~\cite{biswas2021socnavbench,tsoi2022sean,kastner2022arena}, here we follow recent work in HRI~\cite{lo2019perception,mavrogiannis2022social} and focus on the \emph{human-centered} performance of the interactions by evaluating the impact that the controlled robots have on the efficiency and safety of the humans.

\vspace*{-12px}
\subsection{Human Trajectory Prediction}

The need for accurate methods to predict the future movements of humans is crucial to developing better robot simulation and navigation algorithms. To perform pedestrian trajectory prediction, early work has adopted model-based solutions that leverage hand-tuned mathematical models of humans' behavior. These include models based on social forces, velocity-obstacles, and data-driven formulations~\cite{helbing,pellegrini2009you,van2011reciprocal,karamouzas2014universal}. In recent years, model-based methods have been replaced by model-free approaches that rely on deep learning architectures to achieve state-of-the-art (SOTA) prediction performance. Such architectures include RNN structures for generating sequential predictions, social attention and pooling mechanisms for capturing the neighbors' influence, and GAN architectures and conditional VAE (CVAE) models for accounting for the uncertainty and multimodality of human decision making~\cite{alahi2016social,socialattention,gupta2018social,xu2022socialvae, salzmann2020trajectron++}. We refer to the excellent survey of Rudenko et al.~\cite{rudenkosurvey} for more details. Our paper considers two CVAE approaches that have shown to achieve SOTA performance in human-human interaction settings and evaluates their applicability to indoor HRI settings as compared to a model-based baseline.  

\section{Experimental Setup}

\begin{figure}
\vspace{1pt}
  \centering
   \includegraphics[width=\linewidth]{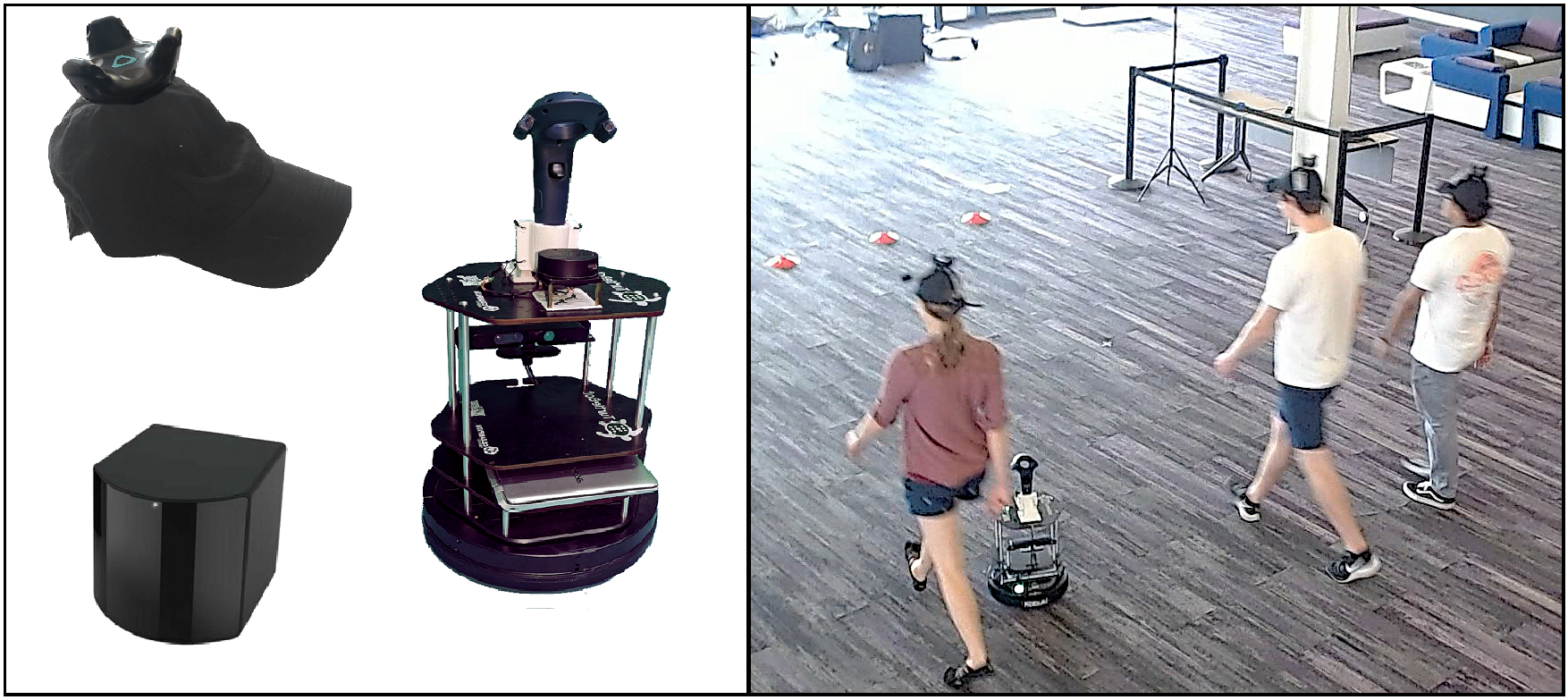}
   \vspace*{-17px}
   \caption{(left) Our tracking solution includes Vive Trackers 2018 attached to hats, a Vive Controller mounted on a Turtlebot, and Vive Base Stations 2.0. (right) A still from the overhead webcam during an experimental trial.
   \vspace{-12px}
   }
  \label{fig:tracking}
\end{figure}

We conducted our experiments in lobby space near our lab measuring approximately  $6\,\text{m} \times 6\,\text{m}$. This space represents a typical environment for the application of indoor service robots. Obstacles visible in Fig.~\ref{fig:tracking} were outside the boundaries of the experimental volume. Cones marked the pedestrian start and goal positions. The participants and robot were given a simultaneous trigger to begin moving, with the participants getting an audible countdown. The Institutional Review Board of Clemson University approved the study on May 4th, 2022 under application number IRB2022-0251. 

\vspace{-12px}
\subsection{Participants}
Two female and four male participants (age between 23 and 30 years old, $\text{M} = 26.6$, $\text{SD} = 1.6$) gave informed consent to participate in this user study. All were free of any known impediments to their walking and had normal or corrected vision, as verified by self-report. The participants were drawn from a pool of graduate students enrolled in the College of Engineering and Applied Sciences at Clemson University. The average height of the participants was $1.75 \pm 0.12 m$, and the shoulder-to-shoulder distance was $0.44 \pm 0.04 m$. None of the students had interacted with robots in their research, but most had been exposed to robotics somehow. To avoid any potential biases, the participants were not informed about the overall goal of the experiments until after the recordings had concluded. During the trials, the participants were instructed to act as they normally do when walking in our lobby space. The experiments were conducted over two consecutive days, with three participants in the first day and three in the second. 

\begin{figure*}
    \vspace{2pt}
    \includegraphics[width=1.\textwidth]{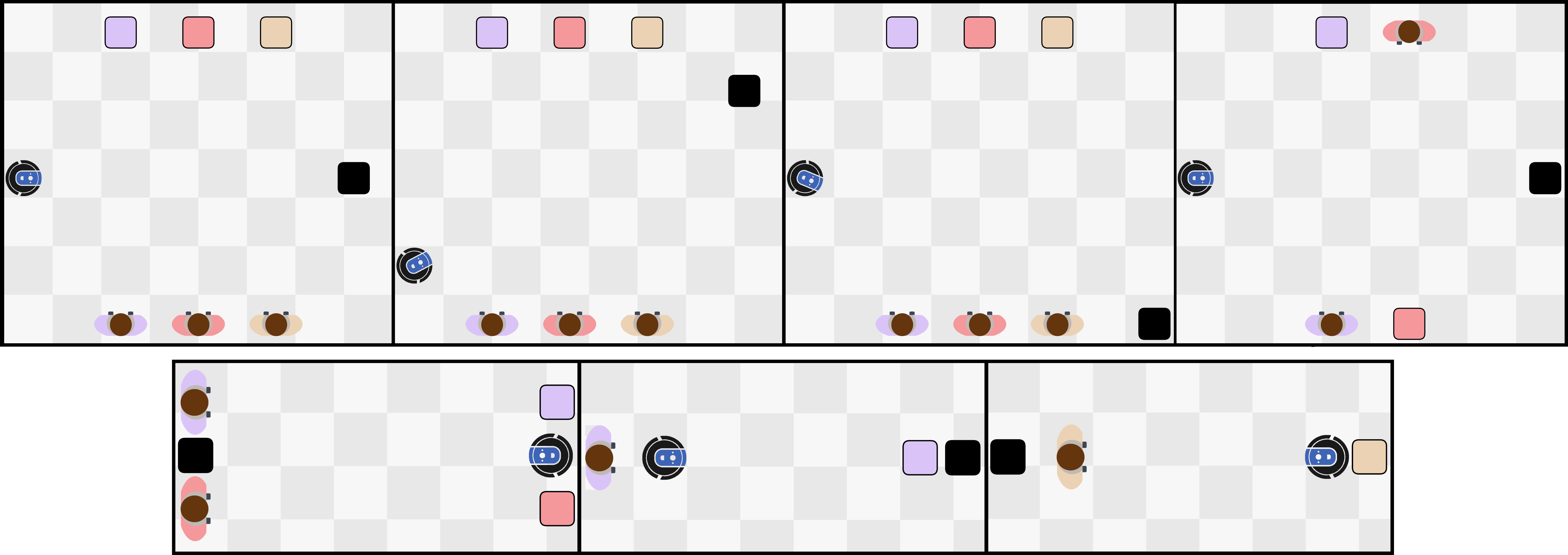}
     \vspace{-15pt}
    \caption{Graphical representation of the interaction scenarios used in our study. The scenarios cover the entire experimental area of 6m $\times$ 6m. From top to bottom and left to right, they are \emph{Perpendicular}, \emph{Adjacent}, \emph{Opposite}, \emph{Intersection}, \emph{2v1}, \emph{Overtake}, and \emph{Head-to-Head}. The top four scenarios were used in the analysis for $H_{1}$ and all seven for $H_{2}$ and $H_{3}$. Colored squares denote the goal locations of the corresponding agents (humans or robot).
    \vspace{-13px}
    }
    \label{fig:scenarios}
\end{figure*}

\vspace*{-7px}
\subsection{Robot}

The wheeled robot the participants interacted with was a TurtleBot 2 from ClearPath Robotics, shown in Fig.~\ref{fig:tracking}. This robot is approximately 42cm tall which may limit the applicability of our study to similarly sized robots \cite{rae2013influence}. To provide a varied dataset, we consider three local planning algorithms to steer the robot to its destination: a \linear controller that maintains a constant speed with no deviation from the straight line path to the robot's goal, NH-TTC~\cite{davis2018nh} which formulates local navigation as an optimization problem employing a time-to-collision based cost function~\cite{karamouzas2014universal}, and CADRL-GAC3~\cite{everett2021collision} which learns a steering policy using reinforcement learning. We chose NH-TTC and CADRL as representative geometric and machine-learning approaches for collision-free local steering, respectively. Both approaches are anticipatory in nature, accounting for potential collisions with nearby neighbors and reacting to them, as opposed to Linear, which is a non-anticipatory approach.  During our experiments, all planning nodes were run on a central server and a velocity command was relayed to the robot, ensuring no delayed robot responses. The robots had a singular goal with no global plan ensuring only the local controller's behavior was analyzed. We used the official Robot Operating System (ROS) implementation of CADRL and a custom ROS wrapper over the official NH-TTC code. In all trials, we set the maximum robot speed at $0.5\,$m/s. The participants were not informed about the possibility of different robot controllers. 

\vspace{-9px}
\subsection{Interaction Scenes}

We define an interaction scene as a combination of start and goal positions for participants and the robot. Figure~\ref{fig:scenarios} shows the seven different scenes considered in our experiments. We refer to the our project website for example videos of the scenes. For each scene, we consider a No Robot baseline and three robot controller cases (Linear, NH-TTC, and CADRL), resulting in four different robot configurations. We define a scenario as a combination of a scene, a choice in robot configuration, and the specific order of the participants. To collect a wide array of interactions, we systematically permuted the order of the participants and the type of the robot controller to ensure that we had a No Robot control scenario for each considered participants' order. Each participant engaged in all seven scenes traversing on average  8.8$\pm$0.2$\,$m in 11.1$\pm$0.8$\,$s. In total, we recorded 97 robot trajectories  and 309 human trajectories capturing 135 different scenarios with a total duration of 25 minutes. 

\vspace{-8px}

\subsection{Motion Tracking System}

All trials were recorded using a tracking system based on the HTC Vive Virtual Reality system (see Fig.~\ref{fig:tracking}). Each participant was given a baseball cap to wear that was modified to mount an HTC Vive Tracker, allowing data collection without the use of Vive's head-mounted display. The robot was tracked with an HTC Vive Controller attached to the top shelf of the Turtlebot. The Vive System interfaced directly with a ROS node, which queried the Vive API at 60Hz to collect the current pose of all devices. The robot’s and humans’ data were obtained via the use of two HTC Lighthouses positioned on either side of the recording area. The accuracy of this system was measured at the sub-centimeter level over a distance of several meters. To reduce oscillations, the recorded robot's and humans' trajectories were resampled at 60Hz, and a Butterworth lowpass filter (2nd order, 0.5Hz cutoff frequency) was applied to the positions. The velocity was estimated from the displacement in the smoothed positions. 

\section{$H_{1}$: Analysis of Robotic Controllers}

As new local planning approaches for mobile robot navigation are being introduced, it is paramount to be able to assess the performance of such approaches. While robot-centric metrics are appropriate in homogeneous environments, in shared environments, the robots should aim to maximize human safety and efficiency even at the cost of their own efficiency. As such, we consider below several human-centric metrics  and propose to evaluate the robot controllers by comparing them to the No Robot case. We limit our analysis to the four scenes shown in the top row of Fig.~\ref{fig:scenarios}. These scenes have sufficient samples to run statistical analysis and involve complex interactions containing all three participants in the environment and/or human-robot interactions at different angles. Given their anticipatory nature, we expect that CADRL and NH-TTC robots will facilitate more optimal behaviors for all humans in the scene. Thus, we hypothesize that: 

\begin{enumerate}
    \item[\textbf{H1:}]
    CADRL and NH-TTC outperform the Linear controller in metrics related to human safety and efficiency when compared against the No Robot case.
\end{enumerate}

\begin{table*}
    \renewcommand{\arraystretch}{1.3}
    \caption{Mean and standard deviation for metrics grouped by scene and controller. Significant differences with respect to the No Robot case are highlighted in \textcolor{ninetyfive}{red} ($p<0.05$). The No Robot case is included in bold. As its statistics are identical for the 3-agent scenes in the top three rows, we only report them for the opposite. 
    }
    \label{table:stats}
    \centering
    \vspace{-5pt}
    \begin{tabularx}{\textwidth}{X|X|X|X|X|X}
        \hline
        \bfseries{Scene} & \bfseries{Robot Case} & \bfseries{Speed (m/s)} & \bfseries{Travel Time (s)} & \bfseries{Path Linearity (m)} & \bfseries{Safety (\%)} \\
        \hline
        \hline
        \multirow{3}{*}{Perpendicular} & Linear & \textcolor{ninetyfive}{0.872 ± 0.384} & \textcolor{ninetynine}{12.121 ± 2.213} & \textcolor{ninetynine}{37.737 ± 14.259} & \textcolor{ninetynine}{0.901 ± 0.1}\\
        {} & CADRL & 1.114 ± 0.094 & 10.868 ± 1.207 & 31.219 ± 16.689 & \textcolor{ninetynine}{0.901 ± 0.11}\\
        {} & NHTTC & 1.085 ± 0.178 & 10.974 ± 1.69 & 32.579 ± 16.264 & \textcolor{ninetynine}{0.866 ± 0.141}\\
        \hline
        \multirow{3}{*}{Adjacent} & Linear & \textcolor{noeffect}{1.082 ± 0.156} & 11.417 ± 1.529 & \textcolor{noeffect}{38.358 ± 29.524} & \textcolor{ninetynine}{0.886 ± 0.117}\\
        {} & CADRL & \textcolor{noeffect}{1.064 ± 0.169} & 11.229 ± 1.697 & \textcolor{noeffect}{33.946 ± 22.505} & \textcolor{ninetynine}{0.878 ± 0.139}\\
        {} & NHTTC & \textcolor{noeffect}{0.942 ± 0.312} & \textcolor{ninetynine}{11.786 ± 1.307} & \textcolor{noeffect}{35.605 ± 26.534} & 0.914 ± 0.112\\
        \hline
        \multirow{4}{*}{Opposite} & Linear & \textcolor{ninetyfive}{1.053 ± 0.073} & \textcolor{ninetynine}{11.383 ± 0.875} & \textcolor{noeffect}{38.885 ± 29.95} & 0.925 ± 0.087\\
        {} & CADRL & \textcolor{ninetynine}{0.993 ± 0.171} & \textcolor{ninetynine}{12.252 ± 1.252} & \textcolor{noeffect}{38.108 ± 27.28} & \textcolor{ninetynine}{0.877 ± 0.105}\\
        {} & NHTTC & \textcolor{ninetynine}{1.012 ± 0.117} & \textcolor{ninetynine}{11.884 ± 1.2} & \textcolor{noeffect}{34.319 ± 28.685} & \textcolor{ninetynine}{0.908 ± 0.094}\\
        {} & \textbf{No Robot} & \textbf{1.134 ± 0.081} & \textbf{10.531 ± 0.509} & \textbf{21.431 ± 8.813} & \textbf{0.992 ± 0.024}\\
        \hline
        \multirow{4}{*}{Intersection} & Linear & 1.079 ± 0.148 & 10.808 ± 1.756 & 45.927 ± 32.669 & \textcolor{noeffect}{0.997 ± 0.011}\\
        {} & CADRL & \textcolor{ninetynine}{1.038 ± 0.094} & \textcolor{ninetynine}{11.1 ± 0.922} & 39.958 ± 22.766 & \textcolor{noeffect}{1.0 ± 0.0}\\
        {} & NHTTC & \textcolor{ninetynine}{0.999 ± 0.159} & \textcolor{ninetyfive}{11.648 ± 1.621} & \textcolor{ninetyfive}{47.102 ± 29.406} & \textcolor{noeffect}{1.0 ± 0.0}\\
        {} & \textbf{No Robot} & \textbf{1.164 ± 0.054} & \textbf{9.535 ± 0.646} & \textbf{20.29 ± 9.381} & \textbf{1.0 ± 0.0}\\
        \hline
    \end{tabularx}
    \vspace{-10pt}
\end{table*}

\vspace{-7pt}
\subsection{Metrics}

Motivated by prior work\cite{zhang2022hri , thorDataset2018, biswas2021socnavbench, mavrogiannis2019effects}, we consider three \emph{efficiency} metrics and a \emph{safety} metric to determine the impact each robot controller has on the participants. These metrics are calculated from the participant's perspective, rewarding human-centered behavior. Formally, given a scenario, let $A$ denote its set of agents that include the set of human participants $H$ and possibly a robot $R$.  Each human participant $H_{i}$ has a start position $\mathbf{S}_{i}$, a goal position $\mathbf{G}_{i}$, and a radius $r_i$ determined by the shoulder-shoulder distance.  Each participant also has an array of valid time steps $T$ and, for each $t \in T$, the position, and velocity of the participant are denoted by $\mathbf{p}_{i}^{t}$ and $\mathbf{v}_{i}^{t}$, respectively. The robot parameters are defined similarly.  We define the following metrics on the human agents:

\subsubsection{Speed}
The speed of each human agent is calculated as the median of the L2 norm for each recorded velocity along the agent's trajectory. This metric was used to judge how unimpeded the participants movement was. The difference from their free walking speed measured at 1.3$\pm$0.1$\,$m/s before our experiments took place should also be taken into account.

\subsubsection{Travel Time} 

The total time a human agent spent traveling between the start and goal positions. This metric was used to judge the overall efficiency of the human paths.

\subsubsection{Path Linearity} 

The path linearity of each human agent is the sum of the absolute deviations from the straight line path from the agent's start to the goal: 
    $\sum_{t \in T}
    \left\vert \dfrac{\lVert (G_i - S_i) \times (S_i - \mathbf{p}_{i}^{t}) \rVert}{\lVert G_i - S_i \rVert} \right\vert$. 
The intention of this metric is not to promote straight paths but rather to ensure that the robot doesn't impact the shape of the participant's path too drastically.

\subsubsection{Safety} 

Let $\tau$ denote the time that it takes for the disks of two agents to collide assuming they both maintain their current velocities. Given that humans have a finite reaction time of ~200-400$\,$ms and an interaction time horizon of 2-4$\,$s~\cite{karamouzas2014universal,olivier2012minimal}, we consider any interaction with a $\tau$-value below $1.5\,$s as potentially dangerous.  The safety of each human agent is then defined as the ratio of the safe 
\vspace{0.15cm}
frames to all recorded trajectory frames: $\dfrac{\left\lvert \left\{ \tau(H_{i}^{t}, A_{j}^{t}) > 1.5, \: \forall t \in T, \: \forall j\neq i \in A \right\} \right\rvert}{\lvert T \rvert}$, 
where $A_j$ denotes either another human participant or the robot, and $\tau(H_i, A_j)$ is a function of the relative velocity, relative displacement, and radii of the two interacting agents. The intention of this metric is to highlight potentially uncomfortable interaction frames that lead to imminent collisions. We  assume that controllers which minimize this metric will impede the movements of the humans. 

\vspace{-10px}
\subsection{Statistics}

To evaluate the effect that the robot has on the participants in each of the four scenes, one-way analysis of variance was performed on each of the proposed measures using the robot type (No Robot, Linear, NH-TTC, CADRL) as independent variable. To guarantee valid comparisons, we performed a Welch ANOVA whenever the assumption of equal variance was violated and the Kruskall-Wallis nonparametric alternative to the classic ANOVA when the normality assumption did not hold~\cite{hsu1996multiple}. Posthoc analysis  revealed no significant differences between the robot controllers in any of the studied metrics and scenes.  To further investigate the human-centered behavior of the robots and determine whether introducing a robot with a specific type of controller (Linear, NH-TTC, CADRL) had a significant impact on the participants, pairwise post-hoc comparisons were performed for each scene-metric combination between each controller and the No Robot case using the latter as the control group. These were done using Dunn's test after nonparametric ANOVAs, and Dunnet's and Dunnett's T3 tests after classic and Welch ANOVAs, respectively~\cite{hsu1996multiple}.

\vspace{-5px}
\subsection{Results}

Regarding the safety on the \emph{Intersection} scene, no statistical comparisons were performed given the lack of variance and large mean indicating minimal unsafe frames. For the rest of the scenes, a significant interaction effect was observed for all metrics, besides the speed on the \emph{Adjacent} and path linearity on both \emph{Adjacent} and \emph{Opposite}. The mean and standard deviation for the metrics as well as the pairwise significance for the post-hoc tests are shown in Table~\ref{table:stats}.  Overall, when focusing on the human navigation efficiency,  none of the controllers is comparable to the No Robot case for the \emph{\opposite} scene, only the \linear controller for \emph{Intersection},  \linear and CADRL for \emph{\adjacent}, and the two anticipatory controllers for \emph{\ninety}. Similarly, in terms of the participants' safety, no controller is universally similar to the No Robot case across all scenes. Therefore, $H_1$ is not supported. 

\vspace{-5px}
\subsection{Discussion}

Our analysis focuses on the impact that a robot controller has on the navigation efficiency and safety of humans. While metrics related to speed and collisions have been used in related work, we use them in a novel way by comparing the performance of the humans with and without the robot in the scene. This provides a more intuitive way to assess the human-centered behavior of different controllers and offers an implicit comparison between them. We also explicitly compared the controllers but observed no significant pairwise differences. However, we argue that even if two controllers are statistically different from each other, if there is no statistical difference from the No Robot case, both controllers should be treated as equally unacceptable regarding human-centeredness. 

In that vein, it is clear from Table~\ref{table:stats} that anticipatory controllers can impede the safety and navigation efficiency of humans and  do not necessarily outperform the non-anticipatory \linear controller. In retrospect, these results can be explained by the \emph{duality} of the HRI problem. In HRI settings, the actions of a robot affect the behavior of the humans as much as the actions of the humans affect the behavior of the robot~\cite{sadigh2016information,trautman2010unfreezing}. So, when an NH-TTC or a CADRL Turtlebot interacts with humans, it tries to predict what the humans will do and react accordingly, sharing the collision avoidance effort. However, such intelligent, anticipatory behavior can sometimes confuse humans and negatively influence their efficiency and safety as they try to guess the robot's behavior. In contrast, when interacting with a Turtlebot that follows a straight path, the humans realize that the robot does not react to them and can quickly make minor adjustments to their trajectories way in advance to resolve collisions efficiently. We note that this may not be true when humans interact in more complex settings where their goals are not in direct sight. However, in our indoor settings, our analysis highlights the need for robot navigation methods that put human safety and efficiency at the forefront along with the importance of evaluating such methods from a human-centric perspective. 

\section{$H_{2}$: Human Trajectory Prediction}

In recent years, with the rise of deep learning techniques, many trajectory prediction approaches have shown impressive results on crowd and vehicle datasets, attributed in part to successfully reasoning about agent-agent and agent-environment interactions~\cite{rudenkosurvey}. While such approaches can find applications to social robot navigation and facilitate more accurate human trajectory prediction, many have not been tested against human-robot-interaction scenarios \cite{mavrogiannis2023core}. To address this issue, we analyze the performance of representative model-free and model-based approaches for human trajectory prediction.  Given short-term historical observations, our goal is to predict the future (ground truth) positions of the participants in our recorded interaction scenarios. A straightforward prediction approach is a constant velocity model (CVM), where the future positions of an agent are inferred from its last position and velocity, assuming a linear extrapolation. In benchmarks commonly used for human trajectory prediction, such as ETH~\cite{pellegrini2009you}, UCY~\cite{lerner2007crowds}, and SDD~\cite{robicquet2020learning}, it has been shown that CVM-based approaches can rival even state-of-the-art neural network approaches~\cite{scholler2020constant}. As such, we use CVM as a baseline and compare its performance to two SOTA approaches for model-free trajectory prediction, Trajectron++~\cite{salzmann2020trajectron++} and SocialVAE~\cite{xu2022socialvae}. In contrast to the deterministic nature of CVM, both Trajectron++ and SocialVAE are stochastic models outputting a future trajectory distribution by exploiting a conditional variational autoencoder architecture. Given the limited amount of complex interactions that humans undergo in our indoor interaction settings, we hypothesize that: 

\begin{enumerate}
    \item[\textbf{H2:}]
    The Constant Velocity Model outperforms SocialVAE and Trajectron++ in terms of prediction error on the Zucker dataset.
\end{enumerate}

\vspace{-5px}
\subsection{Evaluation Dataset and Metrics}

To test $H_2$, we consider all seven scenes of our Zucker dataset. Following the  literature~\cite{alahi2016social,xu2022socialvae,salzmann2020trajectron++}, we resample the recorded robot and human trajectories at 2.5$\,$FPS. We further split the trajectories into segments of $H+T$ frames assuming we have access to $T$ consecutive frames of prior observation. The problem then is to predict the future positions of each human for the next H-frames given the corresponding T-frame observation of the human’s trajectory, where each observation sample includes both the position of the human and the positions of its neighbors. Given the length of the recorded participants' trajectories, we use $T=5$ frames (2$\,$s) as the observation window and $H=8$ frames (3.2$\,$s) as the prediction horizon. In total, we generated three sets of human trajectory prediction data based on the type of the robot controller that the humans interacted with. The combined dataset consists of 42,159 local trajectories that capture a variety of interactions. While the publicly available Trajectron++ and SocialVAE code offers many pre-trained models, we retrain both models to fit our observation horizons. Since the Zucker dataset does not have enough trajectories to facilitate a training-testing split, we retrained the models using the Univ dataset from the UCY benchmark~\cite{lerner2007crowds}. This dataset contains a rich set of agents interacting at varied speeds and angles, including non-reactive and reactive agents, groups, and individuals, sufficiently covering the types of interaction scenarios that humans encountered in our experiments. As in prior work~\cite{gupta2018social,xu2022socialvae}, we use the minimum Average Displacement Error (ADE)  and the minimum Final Displacement Error (FDE) over $k$ predictions to assess a model's performance, where $k\,$=$\,$1 for the deterministic CVM and $k\,$=$\,$5 for the multimodal Trajecton++ and SocialVAE models. ADE measures the average $L2$ distance in meters between the predicted positions and the ground truth ones over the whole predicted trajectory while FDE only focuses on the distance at the end of the predicted trajectory. 

\begin{table}[t]
    \renewcommand{\arraystretch}{1.2}
    \caption{
    ADE/FDE in meters of CVM, Trajectron++, and SocialVAE in Zucker scenes grouped by robot controllers. We report the best of 5 predictions for Trajectron++ and SocialVAE.     
    }
    \vspace{-5pt}
    \label{tab:zucker_traj}
    \centering
    \begin{tabular}{c|c|c|c|c}
        \hline
        \bfseries Method & \bfseries CADRL   & \bfseries NH-TTC  & \bfseries Linear   & \bfseries Average  \\
        \hline
        \hline
        CVM       & 0.21 / 0.42 & 0.19 / 0.38 & 0.20 / 0.41 & 0.20 / 0.40 \\
        Trajectron++  & 0.17 / 0.33 & 0.16 / 0.32 & 0.14 / 0.28 & 0.16 / 0.31 \\
        SocialVAE   & 0.16 / 0.32 & 0.15 / 0.29 & 0.15 / 0.30 & 0.15 / 0.30 \\   
        \hline
    \end{tabular}
    \vspace{-9pt}
\end{table}

\vspace{-10px}
\subsection{Results}

In Table~\ref{tab:zucker_traj}, we report the performance of CVM, Trajectron++, and SocialVAE on the three robot scenes of our Zucker evaluation dataset. Overall, the performance of the CVM baseline can provide a quick way to assess the interaction complexity in the scene~\cite{scholler2020constant}. Small ADE/FDE values typically denote that humans exhibit mostly linear interactions, with the past trajectory of a tracked human being the main feature for predicting its future. Given the CVM performance, one could argue that humans do not undergo too complex social interactions in Zucker. Therefore, employing a linear model to track the trajectories of humans is a strong baseline. However, as seen in Table~\ref{tab:zucker_traj}, the model-free approaches attain lower prediction errors, outperforming the CVM baseline across the different scenes. As such, we reject $H_2$. 

\vspace{-10px}
\subsection{Discussion}

\begin{figure}[t]
    \centering
    \includegraphics[width=1.09\columnwidth]{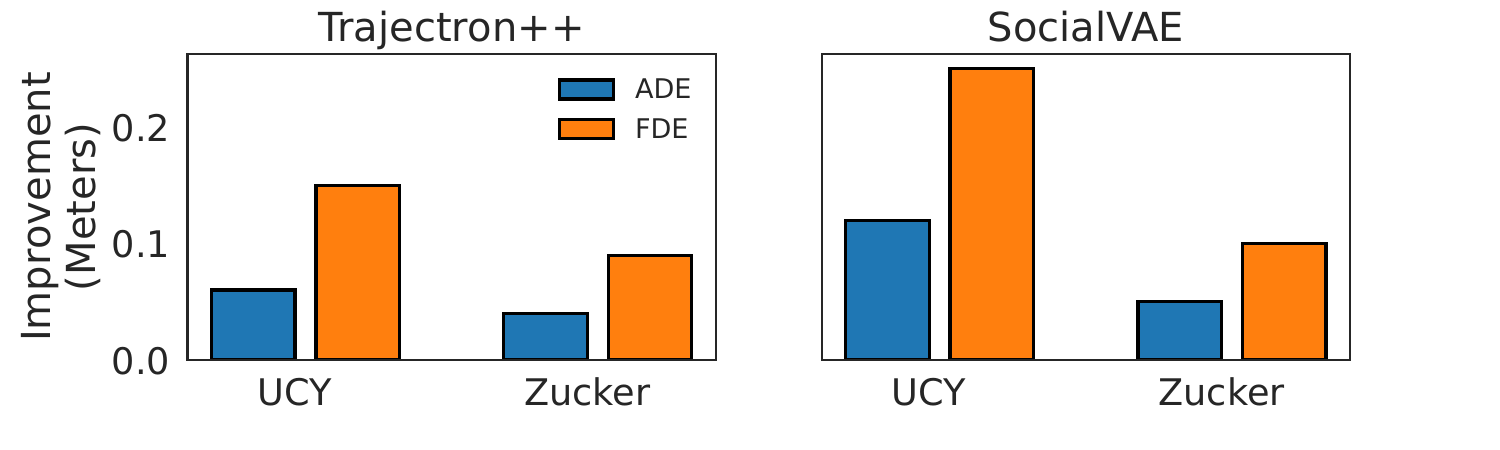}
    \vspace*{-25pt}
    \caption{ADE/FDE from Trajectron++ and SocialVAE on the UCY and Zucker datasets in terms of improvement over the CVM. The Zucker results are obtained using all three robot steering controllers.}
    \label{fig:traj_pi}
    \vspace*{-1em}
\end{figure}

Our analysis shows that model-free approaches such as Trajectron++ and SocialVAE can provide reliable human trajectory predictions in small indoor settings that can potentially inform more human-aware robot control. On the other hand, a linear model can serve as a strong baseline in similar HRI indoor settings, especially as more complex methods might introduce unnecessary computational overhead in an already strained real-time setting. To put the above results into perspective, we also compute the performance of the three studied trajectory prediction approaches on the pedestrian datasets from the UCY benchmark. In Fig.~\ref{fig:traj_pi}, we report the  performance improvement of Trajectron++ and SocialVAE over CVM for both the UCY and Zucker datasets. Similar to Zucker, the model-free approaches outperform CVM. However, the performance improvement is much lower in our Zucker dataset than in the human-only UCY datasets commonly used in trajectory prediction benchmarks. This indicates that there is potential room to improve the performance of human trajectory prediction methods in indoor HRI settings by, e.g.,  training models with more representative  datasets, devising new models, and/or refining trained models both in online and lifelong manners~\cite{moracontinual}.

\vspace{-4px}
\section{$H_{3}$: Human-Robot Interactions}

In~\cite{karamouzas2014universal} it was shown that the  ``interaction energy" between pairs of humans in a crowd can be modeled through an inverse power-law relationship with their projected time to collision, $\tau$. The intuition is that the sooner two pedestrians are going  to collide, the more uncomfortable they feel leading to a high energy to avert the collision. This mathematical model of human-human interactions is therefore anticipatory, depending not just on the current state of the environment but also on expected future states. While its applicability has been validated in many different conditions and settings, it was derived from human-only datasets. Here we are interested in gaining a better understanding  of heterogeneous interactions between humans and small service robots in indoor spaces like the ones considered in our user study. We believe that, in such settings, humans will be more cautious and try to avoid collisions with the robot as early as possible. As such, we hypothesize that:

\begin{enumerate}
    \item[\textbf{H3:}]  
    The interaction energy between human-robot pairs in the Zucker dataset is different than that observed between human-human pairs in crowd-only datasets.
\end{enumerate}

\vspace{-5px}
\subsection{Methodology}

Following~\cite{karamouzas2014universal}, we focus on the time to collision metric, $\tau$, and employ a probabilistic analysis approach to study pairwise interactions in isolation from all other factors that can influence the behavior of agents. To do so, we compare the frequency of $\tau$-values between human-robot pairs that appear in a recorded scene at the same time to the frequency of $\tau$-values for random, non-interacting pairs. The resulting normalized frequency estimates the pair distribution function, $g(\tau)$, that highlights statistically suppressed interactions. Such a function can be converted to a ``social interaction energy", $E$, assuming a Boltzmann-like relation between $g$ and $E$: $E(\tau) \propto log(1/g(\tau))$. Thus, the energy vanishes when $g(\tau)=1$ and drastically increases for small $g(\tau)$ values. 

\vspace{-5px}
\subsection{Results}
\begin{figure}
    \centering
    \includegraphics[width=1.0\columnwidth]{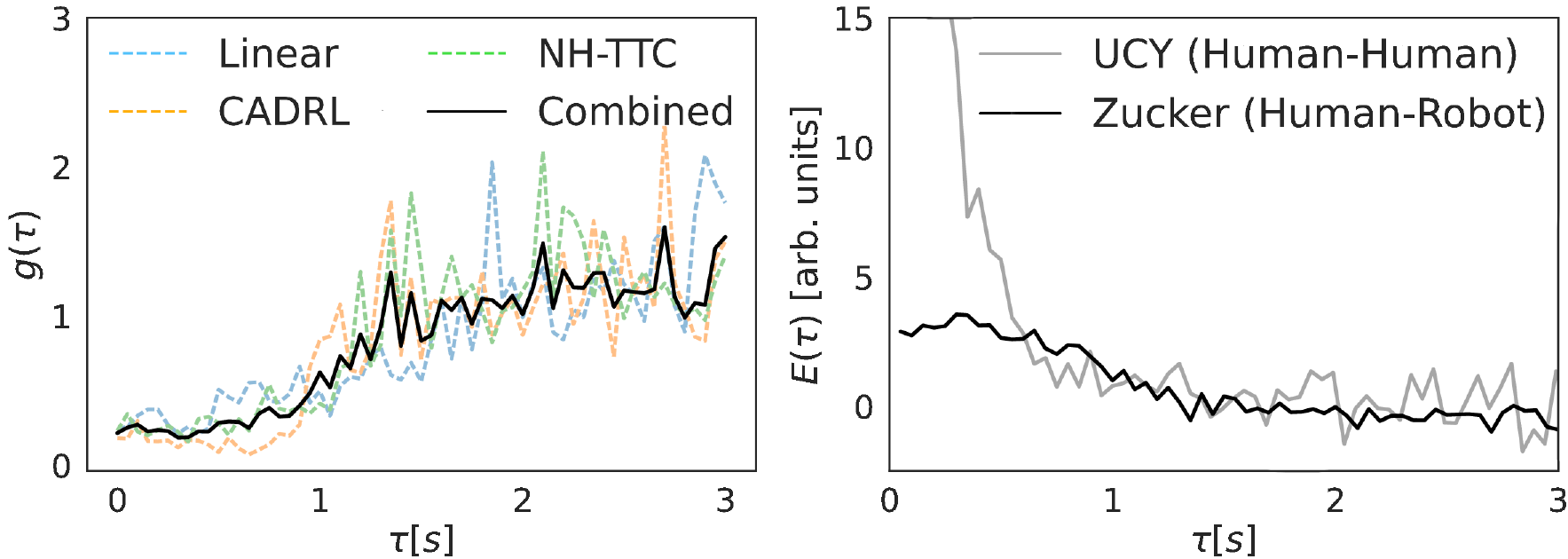}   
    \vspace*{-20pt}
    \caption{(left) The pair distribution function $g$ as a function of the projected time-to-collision value $\tau$ between human-robot pairs. The curves are grouped based on the type of controller that the robot employs and exhibit similar behavior, allowing us to combine them. (right) The corresponding interaction energy $E$ remains finite even for small $\tau$-values, as opposed to the human-human interaction energy inferred from the UCY crowd dataset (obtained from~\cite{karamouzas2014universal}). The energy curves are normalized so that $E(1) = 1$.
    \vspace*{-1em}
    }
    \label{fig:robot_law}
\end{figure}

Figure~\ref{fig:robot_law} left shows the pair distribution function $g(\tau)$ from the Zucker dataset. The reported plots were obtained by considering $\tau$-values for human-robot-only pairs. Overall, the NH-TTC, CADRL, and Linear curves show very similar behavior, allowing us to combine them and estimate the normalized distribution of the $\tau$-values between the human participants and robot controllers. We note that during our HRI experiments, no actual collisions took place, though several near collisions were observed (the positive  $g(\tau)$ value for $\tau=0$ is due to the binning of $\tau$). In Fig.~\ref{fig:robot_law} right, we show the interaction energy $E(\tau)$ derived from $g(\tau)$. We also depict the interaction energy graph from~\cite{karamouzas2014universal} obtained by analyzing pedestrian interactions in the UCY crowd scenes. As can be seen, the Zucker and UCY energy plots are visibly different. A two-sample Kolmogorov-Smirnov test statistically confirmed that the respective $E(\tau)$ values are coming from different distributions. Thus, we accept $H_3$.

\vspace*{-10px}
\subsection{Discussion}

Our results show that in both our HRI scenes and the crowd interaction scenes, humans don't care about collisions that will take place in the far future ($E(\tau)\approx 0$ for $\tau>1.5\,$s in both UCY and Zucker). Furthermore, humans anticipate collisions when interacting with other agents (humans or Turtlebots), with the energy increasing with decreasing time-to-collision values. However, when faced with an imminent collision, and in contrast to what we expected before the experiments, the interaction energy between a human and a Turtlebot is finite as opposed to human-human interactions that exhibit infinite energy. Specifically, $E(\tau)$ rises to infinity when $\tau< 0.3\,$ for the UCY pairs, while it saturates around a constant value for such $\tau$'s in Zucker. In hindsight, such a finding is not that surprising. Turtlebots can only cause a finite amount of social discomfort, as they move slowly and are not intimidating. For example, we typically step over Roombas, highlighting our lack of any fear of colliding with them. Hence, humans are more willing to head toward a collision since they can resolve it quickly. We expect the observed interactions to extend to other small indoor robots leading to the development of more accurate models of human behavior compared to the ones currently employed in robot simulators~\cite{helbing,van2011reciprocal,karamouzas2014universal, kastner2022arena}. We acknowledge, though, that such finite interaction forces may not apply to larger indoor robots such as humanoids or fast-moving transport robots deployed in fulfillment centers. 

\vspace{-8pt}
\section{Conclusion}

In this paper, we study how human behavior is affected by different types of robot controllers in indoor navigation settings. We conclude by highlighting some of the limitations and potential applications of our user study. 

\noindent\emph{Limitations.}
Our work focuses on sparse interactions between humans and small differential drive robots, and thus the conclusions drawn from our study do not necessarily translate to other settings or types of robots. Although, we believe that the interactions are representative of ones that humans could experience in their households and workspaces. Our experiments took place in the lobby area of a building providing a more real-world setting than a traditional laboratory space. However, the participants were aware that they were involved in a user study, which could influence their behavior. Finally, we note that our user study includes a small number of participants which can limit the generalization of our results. Still, our analysis can be broadened with further data collection, an avenue we aim to explore in subsequent studies. 

\noindent\emph{Applications and Future Work.} Despite the aforementioned limitations, we believe that our work  can facilitate future research in \emph{social robot navigation}, \emph{human trajectory prediction}, and \emph{crowd simulation}.  In particular, regarding \emph{social robot navigation}, we propose an intuitive way to measure the \emph{human-centered} behavior of a robot by comparing the performance of humans with and without the robot. The related analysis shows the need to invest in more human-centered methods for mobile indoor navigation~\cite{vassallo2018walkers}, seeking to prioritize human over robot performance (e.g., for tasks like vacuuming that aren't constrained by time). Our evaluation approach could also be applied to recent work that has explored group-aware robot planning techniques and collaborative strategies for robot collision avoidance, among others~\cite{sun2021move,mavrogiannis2022winding,katyal2022learning,gonon2023inverse}.

Regarding \emph{human trajectory prediction},   the performance results reported in Table~\ref{tab:zucker_traj} can serve as a strong baseline for testing other prediction methods. We further release our  evaluation HRI  dataset along with our pretrained models to facilitate the refinement of existing trajectory prediction methods and the development of new ones~\cite{rudenkosurvey}. Similarly, our dataset can find applications to continual learning methods that can help improve existing trajectory prediction models in settings with continuously adapting human behavior~\cite{moracontinual}. Finally, our dataset can inform the development of  more accurate \emph{crowd simulation} models that can facilitate the training and evaluation of robot navigation methods.  In existing robot simulators and benchmarking tools~\cite{everett2021collision,chen2019crowd,tsoi2022sean,kastner2022arena}, human behavior is typically modelled with distance-based potentials~\cite{helbing} or anticipatory ones based on the notion of time to collision, $\tau$~\cite{van2011reciprocal,karamouzas2014universal}. However, our analysis suggests a combination of the two may be a more accurate model for human-robot interaction.  As such, the corresponding $E(\tau$) plot  in Fig.~\ref{fig:robot_law} can be used to derive a geometric model for human-robot interactions.  Alternatively,  our collected human trajectories can serve as expert demonstrations to train data-driven crowd models through reinforcement learning~\cite{kim2016socially,kretzschmar2016socially}.

\vspace*{-6pt}

\bibliographystyle{IEEEtran}
\bibliography{main}

%% file: appendix.tex
\renewcommand{\thefigure}{S\arabic{figure}}
\renewcommand{\thetable}{S\arabic{table}}
\renewcommand{\theequation}{S\arabic{equation}}

\setcounter{figure}{0}
\setcounter{table}{0}
\setcounter{equation}{0}
\setcounter{section}{0}

\renewcommand{\theHfigure}{Supplement.\thefigure}
\renewcommand{\theHtable}{Supplement.\thetable}
\renewcommand{\theHequation}{Supplement.\theequation}

\title{\LARGE A Study in Zucker: Insights on Human-Robot Interactions\\  Supplementary Material}

\author{Alex Day and Ioannis Karamouzas}

\maketitle

\section{Experimental Setup}
\subsection{Zucker Dataset Information}
Table~\ref{table:simple_stats} describes the number of trajectories and information for all participants in our HRI study. \newtext{Table~\ref{table:simple_robot_stats} shows these same statistics grouped by robot controller.}

In Table~\ref{table:datasets}, we also show how the resulting Zucker dataset compares to other similar datasets in terms of experimental trials, overall recording time, robot controllers, and type of interactions. 
\begin{table}[h!]
    \renewcommand{\arraystretch}{1.3}
    \begin{center}
    \caption{Comparison of related human-robot interaction datasets}
    \label{table:datasets}
    \centering
    \begin{tabular}{c|c|c|c}
        \hline
        \bfseries Name & \bfseries \# Trials & \bfseries Time (hh:mm) & \bfseries Type \\
        \hline\hline
        TH\"{O}R        & 13  & 00:60 & Structured\\
        JRDB            & 54  & 01:04 & Unstructured\\
        FLOBOT          & 6   & 00:27 & Unstructured\\
        SCAND           & 138 & 08:42 & Unstructured\\
        \textbf{Zucker} & \newtext{135} & \newtext{00:25} & Structured\\
        \hline
    \end{tabular}
    \end{center}
\end{table}

\begin{table*}[ht]
    \renewcommand{\arraystretch}{1.3}
    \caption{Participant information}
    \label{table:simple_stats}
    \centering
    \begin{tabular}{c|c|c|c|c|c|c}
        \hline
        \bfseries Index & \bfseries \# Trajectories & \bfseries Average Time (s) & \bfseries Average Dist (m) & \bfseries Height (m) & \bfseries Radius (m) & \bfseries Gender \\ 
        \hline\hline
        3          & \newtext{55} & \newtext{10.3 $\pm$ 1.4} & \newtext{8.6 $\pm$ 0.7} & 1.91 & 0.23 & Female  \\
        4          & \newtext{55} & \newtext{10.2 $\pm$ 1.3} & \newtext{8.6 $\pm$ 0.7} & 1.75 & 0.24 & Male  \\
        5          & \newtext{50} & \newtext{10.5 $\pm$ 1.3} & \newtext{8.6 $\pm$ 0.9} & 1.63 & 0.2 & Male  \\
        6          & \newtext{52} & \newtext{12.0 $\pm$ 1.3} & \newtext{8.9 $\pm$ 0.5} & 1.63 & 0.2 & Female \\
        7          & \newtext{46} & \newtext{11.8 $\pm$ 1.5} & \newtext{9.0 $\pm$ 0.6} & 1.88 & 0.24 & Male  \\
        8          & \newtext{51} & \newtext{11.6 $\pm$ 1.2} & \newtext{9.0 $\pm$ 0.4} & 1.73 & 0.22 & Male  \\
        \hline
    \end{tabular}
\end{table*}

\begin{table*}
    \renewcommand{\arraystretch}{1.3}
    \caption{\newtext{Robot information}}
    \label{table:simple_robot_stats}
    \centering
    \begin{tabular}{c|c|c|c|c|c}
        \hline
        \newtext{\bfseries Controller} & \newtext{\bfseries \# Trajectories} & \newtext{\bfseries Average Time (s)} & \newtext{\bfseries Average Dist (m)} & \newtext{\bfseries Height (m)} & \newtext{\bfseries Radius (m)}  \\ 
        \hline\hline
        \newtext{CnewtextRL} &  \newtext{34} & \newtext{10.8 $\pm$ 1.4} & \newtext{3.8 $\pm$ 0.6} & \newtext{0.41} & \newtext{0.18}\\
        \newtext{NHTTC}  & \newtext{32} & \newtext{11.0 $\pm$ 1.4} & \newtext{3.8 $\pm$ 0.6} & \newtext{0.41} & \newtext{0.18}\\
        \newtext{Linear} & \newtext{34} & \newtext{11.1 $\pm$ 1.7} & \newtext{3.8 $\pm$ 0.6} & \newtext{0.41} & \newtext{0.18}\\
        \hline
    \end{tabular}
\end{table*}

\subsection{Interaction Scenes}

\noindent The following four scenes were used in \newtext{$H_{1}$, $H_{2}$, and $H_{3}$}:
\begin{enumerate}
    \item \textbf{Perpendicular: }
    Three pedestrians start in a line-abreast formation and have to walk to their goals directly ahead of them while crossing paths perpendicularly with a robot.

    \item \textbf{Adjacent: }
    Three pedestrians start in a line-abreast formation and have to walk to their goals directly ahead of them while crossing paths with a robot at an acute angle.
    
    \item \textbf{Opposite: } 
    Three pedestrians start in a line-abreast formation and have to walk to their goals directly ahead of them while crossing paths with a robot at an obtuse angle.

    \item \textbf{Intersection: } Two pedestrians stand opposite each other with the goal of swapping their positions. A small offset is added to avoid symmetrical patterns. The robot and its goal are placed on a line orthogonal to the pedestrian's trajectories. 
\end{enumerate}

\vspace{5pt}
\noindent The following scenes were used only in \newtext{$H_{2}$ and $H_{3}$}:
\begin{enumerate}
    \item \textbf{Overtake: } The robot starts ahead of the pedestrian and their goals lie on the same line. 
    \item \textbf{Head-to-Head: } 
    A robot and a pedestrian cross paths while approaching from opposite sides of the environment. The robot's goal is behind the pedestrian's starting position and vice versa.
    \item \textbf{2v1: } 
    A group of two pedestrians has to swap positions with a standalone robot. The pedestrian goals lie on either side of the robot's starting position, and the robot's goal is between the pedestrians' start positions.
\end{enumerate}

\section{\newtext{$H_{1}$: Analysis of Robot Controllers}}
\subsection{\newtext{Statistical Results}}

\newtext{In Table~I we show the means and standard deviations of the metrics. Below we describe the details of the results of our analysis including the p-values.}

\subsubsection{\newtext{Speed}}

\newtext{Our results indicate a significant interaction effect for speed in all but the \emph{\adjacent} scene. By comparing the No Robot to the robot controller cases, we observe that the participant's speed in the \emph{\ninety} scenarios is significantly reduced only when a \linear controller is employed ($p<0.05$). In contrast, in the \emph{Intersection} scenarios, the 
two anticipatory controllers (CADRL, NH-TTC) lead to a significant decrease in speed ($p$-values $<0.05$) as opposed to the \linear controlled robot that does not impede the speed of the participants ($p=0.23$). Finally, in \emph{\opposite}, a significant difference is observed between the No Robot and the robot cases for each of the three controllers ($p<0.01$ for NHTTC, $p<0.05$ for \linear and CADRL).}

\subsubsection{\newtext{Travel Time}}
\newtext{Overall, there is a significant interaction effect for travel time in all scenes. Regarding the 3-participants scenarios, in the \emph{\opposite}, humans needed significantly more time to reach their goals regardless of the type of controller employed ($p$-values $<0.01$), while in \emph{\ninety}, only the linear controller leads to an increase in travel time ($p<0.05$) and only NH-TTC in the \emph{\adjacent} ($p<\,$0.01). In the \emph{Intersection}, a significant difference between the No Robot and robot cases is observed only when humans interact with an anticipatory  robot ($p<0.01$).}

\subsubsection{\newtext{Path Linearity}}
\newtext{In terms of path linearity, no interaction effect was observed in the \emph{\adjacent} and \emph{\opposite} scenarios. As shown in Table~I, the path linearity is significantly different from the No Robot case when humans interact with a \linear controlled robot in the \emph{\ninety} scenarios ($p<0.01$). In \emph{Intersection}, only the NH-TTC controller significantly increases path linearity ($p<0.05$).}

\subsubsection{\newtext{Safety}}
\newtext{A significant interaction effect for safety was observed in all scenes besides the \emph{Intersection}. In \emph{Intersection}, no statistical comparisons were performed given the lack of variance and large mean indicating minimal unsafe frames ($M=0.999$ and $SD=0.006$). In \emph{\ninety}, the human safety is reduced when the robot is present independent of the controller type \mbox{($p$-values$\,<0.05$)}. In \emph{\adjacent}, CADRL and \linear lead to a decrease in safety ($p$-values $<0.01$), and the same applies to the  CADRL and NH-TTC controllers in the \emph{\opposite} scenarios ($p<0.01$ and $p<0.05$, respectively).}

\subsection{\newtext{Minimum Predicted Distance}}
\begin{figure}
\begin{minipage}{.50\columnwidth}
  \centering
   \includegraphics[width=1.\columnwidth]{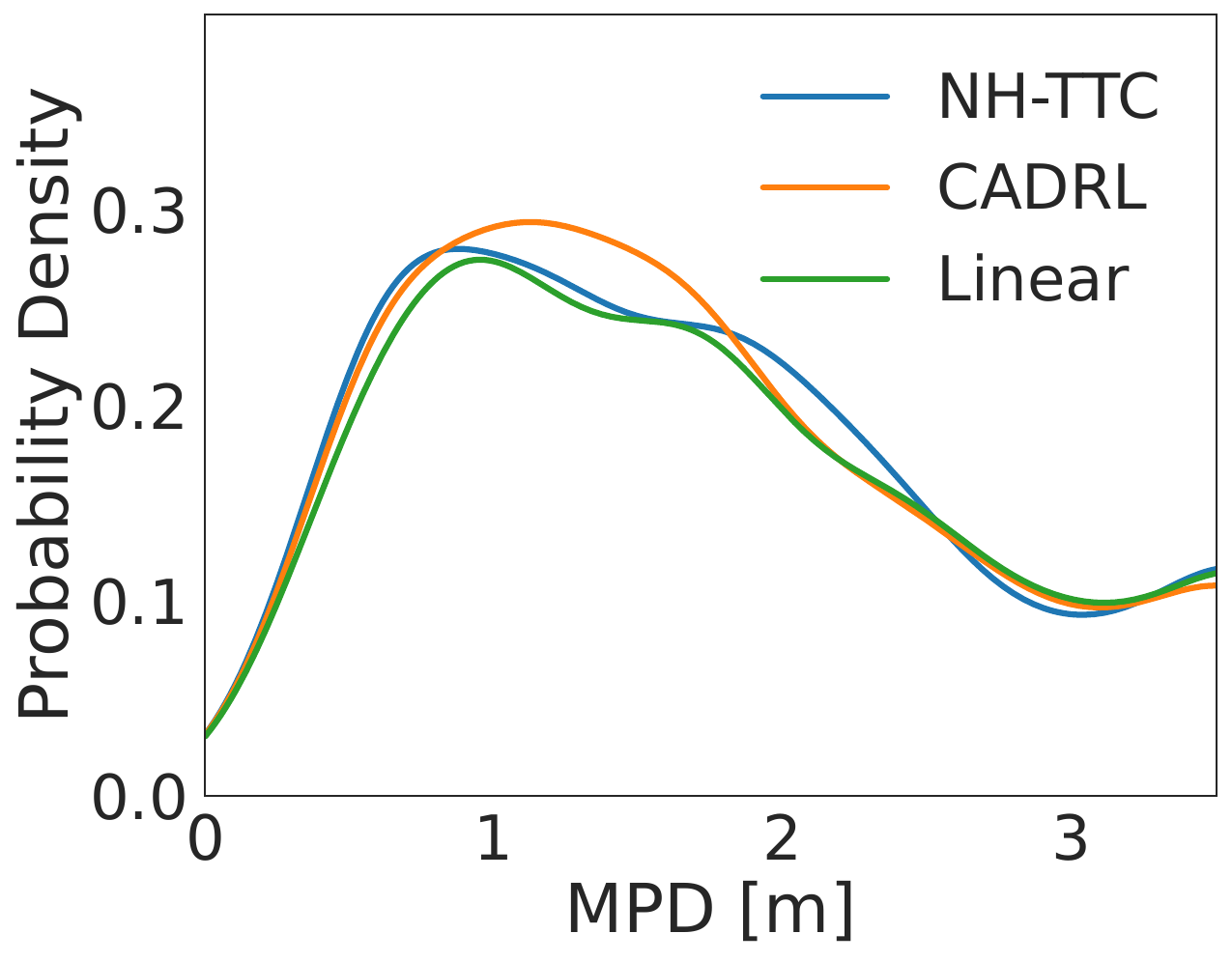}
\end{minipage}%
\begin{minipage}{.50\columnwidth}
\vspace{-15pt}
\hspace{10pt}
\centering
\footnotesize
 \renewcommand{\arraystretch}{1.15}
 \begin{tabular}{l|l|l}
        \hline
        \multicolumn{3}{c}{Jensen-Shannon Distance}\\
        \hline \hline
        Linear & CADRL  & 0.074  \\
        CADRL  & NH-TTC & 0.072  \\
        NH-TTC & Linear & 0.073  \\
        \hline 
    \end{tabular}
\end{minipage}
\vspace{-9pt}
\caption{\newtext{(left) Distribution of the minimal predicted distance (MPD) between a human participant and the Turtlebot clustered by the type of the robot controller employed.The corresponding plots were obtained by considering the center-to-center MPD values of all distinct human-robot interaction pairs. (right) The Jensen-Shannon distance between the MPD distributions.}}
\label{fig:mpd}
\vspace{-2pt}
\end{figure}

\newtext{Minimum Predicted Distance (MPD) allows us to quantify the actual and predicted amount of personal space humans maintain when interacting with the different types of robots.  For example, if a controller has an MPD distribution with peaks inside the personal space of humans (typically between 0.76 to 1.22 m\cite{hall1966hidden}), we can categorize it differently than a controller with peaks further away.}

\newtext{To gain a better understanding of the humans' safety in presence of the robot, in Fig.~\ref{fig:mpd}  we report the distribution of the MPD between human-robot pairs for each of the three robot controllers.  Here, we compute the MPD as the closest center-to-center distance between the two interacting parties assuming a linear extrapolation of their current velocities.  Formally, given two agents $A_i$ and $A_j$, located at $\mathbf{p}_i$ and $\mathbf{p}_j$, respectively, and having velocities $\mathbf{v}_i$ and $\mathbf{v}_j$, the MPD is defined as in~\cite{olivier2012minimal}:}

\begin{equation}
    \label{eq:mpd}
     \min_{t \geq 0} \lVert (\mathbf{x}_{i} - \mathbf{x}_{j}) + (\mathbf{v}_{i} - \mathbf{v}_{i})\,t \rVert.
\end{equation}

\newtext{As shown in the figure, all three robot controllers exhibit similar MPD distributions both qualitatively and quantitatively.}

% \section{$H_{3}$: Human Trajectory Prediction}

% Table~\ref{tab:traj} shows the results of the various trajectory prediction methods on the testing subsets of the UCY dataset~\cite{lerner2007crowds}. Both Trajectron++~\cite{salzmann2020trajectron++} and SocialVAE~\cite{xu2022socialvae} models were trained and tested with 5 frames of history and 8 frames of prediction to adhere to the results reported in Table~III of the main text. Given the multimodal predictions of both models, we report the minimum ADE/FDE over the top-5 predictions,   with SocialVAE using a Final Position Clustering technique in post-processing to help reduce sampling bias~\cite{xu2022socialvae}. We refer to the official repositories of Trajectron++ and SocialVAE for additional details. CVM produces determinstic predictions assuming that the agents maintain their last observed velocities for 8 consecutive frames. 

% \balance

\section{\newtext{$H_{2}$: Human Trajectory Prediction}}

\subsection{UCY Results}
\newtext{In Table~\ref{tab:traj_ucy} we show the results from our retrained models on the UCY datasets.}

\begin{table}[h!]
    \renewcommand{\arraystretch}{1.3}
    \caption{ADE/FDE performance of CVM, Trajectron++, and SocialVAE in the UCY crowd datasets measured in meters.}
    
    \label{tab:traj_ucy}
    \centering
    \begin{tabular}{c|c|c|c|c}
        \hline
        \bfseries Method & \bfseries Zara01 & \bfseries Zara02 & \bfseries Univ & \bfseries \newtext{Average} \\
        \hline
        \hline
        CVM          & 0.26 / 0.55 & 0.22 / 0.47 & 0.32 / 0.68 & \newtext{0.27 / 0.57} \\
        Trajectron++ & 0.20 / 0.40 & 0.17 / 0.34 & 0.26 / 0.53 & \newtext{0.21 / 0.42} \\
        SocialVAE    & 0.17 / 0.35 & 0.14 / 0.28 & 0.13 / 0.23 & \newtext{0.15 / 0.32}\\   
        \hline
    \end{tabular}
\end{table}

\bibliography{main}
\bibliographystyle{IEEEtran}